\documentclass[12pt,a4paper]{article}
\usepackage{authblk}
\usepackage{amsmath,amssymb}
\usepackage{amsthm}
\usepackage{graphicx} 
\usepackage{float}
\usepackage[backend=biber,style=authoryear]{biblatex}
\title{APC-GNN++: An Adaptive Patient-Centric GNN with Context-Aware Attention and \ Mini-Graph Explainability for Diabetes Classification\\[1.5ex]

\large December 20, 2025}
\author{Berkani Khaled}
\affil{bberkani109@gmail.com \\
 \texttt{University of Batna 2, Algeria}}
\date{}

\begin{document}

\maketitle

\section*{ Abstract}

We propose APC-GNN++, an adaptive patient-centric Graph Neural Network for diabetes classification. Our model integrates context-aware edge attention, confidence-guided blending of node features and graph representations, and neighborhood consistency regularization to better capture clinically meaningful relationships between patients. To handle unseen patients, we introduce a mini-graph approach that leverages the nearest neighbors of the new patient, enabling real-time explainable predictions without retraining the global model. We evaluate APC-GNN++ on a real-world diabetes dataset collected from a regional hospital in Algeria and show that it outperforms traditional machine learning models (MLP, Random Forest, XGBoost) and a vanilla GCN, achieving higher test accuracy and macro F1-score. The analysis of node-level confidence scores further reveals how the model balances self-information and graph-based evidence across different patient groups, providing interpretable patient-centric insights. The system is also embedded in a Tkinter-based graphical user interface (GUI) for interactive use by healthcare professionals .\vspace{0.5cm} 

\textbf{Keywords:} Graph Neural Networks; Diabetes classification; Patient similarity graphs; Attention mechanisms; Confidence estimation; Adaptive k-NN; Regularization; Explainable AI; Clinical decision support.

\section{Introduction}

Diabetes classification is a crucial task for personalized treatment and early intervention, as misdiagnosis or delayed diagnosis can lead to severe health complications. Traditional machine learning models often treat patient data independently, ignoring the rich relational structure that may exist between patients with similar clinical profiles. While models such as Random Forests, XGBoost, and multi-layer perceptrons have shown reasonable performance, they lack mechanisms to exploit inter-patient relationships and provide explainable predictions.

Graph Neural Networks (GNNs) have emerged as a powerful tool to model structured data, allowing nodes to exchange information through graph edges. In healthcare, patient similarity graphs can represent patients as nodes connected by clinically meaningful relationships. However, standard GNNs face several limitations: they often assume a fixed graph structure, do not dynamically assess node-level confidence, and fail to provide interpretable explanations for individual predictions [5].

In this work, an Adaptive Patient-Centric Graph Neural Network, referred to as APC-GNN++, is introduced for diabetes classification. The proposed model combines several key innovations:

\begin{itemize}
    \item \textbf{Context-Aware Edge Attention:} assigns dynamic weights to edges based on node features, capturing the relevance of inter-patient relationships.
    \item \textbf{Confidence-Guided Blending:} integrates node-level confidence scores to adaptively combine graph-based embeddings with original patient features.
    \item \textbf{Neighborhood Consistency Regularization:} encourages smooth embeddings among connected nodes, improving generalization.
    \item \textbf{Mini-Graph Explainability for New Patients:} constructs a local $k$-NN graph for unseen cases, enabling interpretable predictions without retraining the global model.
\end{itemize}

By integrating these components, APC-GNN++ improves classification accuracy compared to baseline models while providing interpretable, patient-centric predictions suited to clinical decision support. Moreover, the system is embedded in a user-friendly Tkinter-based graphical interface, facilitating interactive usage by healthcare professionals in real-world settings.

The remainder of this work is structured as follows. First, related research on graph neural networks and diabetes prediction is reviewed. Next, the APC-GNN++ architecture and the adopted training methodology are described in detail. Then, experimental results and an in-depth analysis of performance and interpretability are presented. Finally, the study is concluded with a summary of the main findings and a discussion of promising directions for future research.

\section{Related Work}
\label{sec:related_work}

Graph Neural Networks (GNNs) have become increasingly popular for modeling structured data in healthcare and biomedical applications. Early works such as Graph Convolutional Networks (GCNs)  and Graph Attention Networks (GATs) demonstrated that message passing over graphs can capture complex relational information, leading to improved predictive performance over traditional feature-based models.

In the medical domain, GNNs have been applied to patient similarity graphs for tasks such as disease prediction , patient outcome modeling , and electronic health record analysis . These approaches exploit the relational structure between patients, allowing information from similar patients to enhance individual predictions. However, most existing models assume a fixed graph structure and do not provide mechanisms for dynamically assessing node-level confidence or feature blending, which are crucial for heterogeneous clinical datasets [3].

Explainable AI (XAI) methods have been proposed to increase trust in healthcare models. Techniques such as SHAP  and GNNExplainer  aim to provide interpretability for feature importance and edge contributions. Nonetheless, these methods typically require post-hoc analysis and do not integrate interpretability directly into the prediction process.

For diabetes classification specifically, traditional machine learning models—including Random Forests , XGBoost , and multi-layer perceptrons (MLPs)—operate on independent patient vectors, ignoring relational information. Recent works have started exploring graph-based approaches for patient similarity, but they often lack confidence-guided feature integration and mechanisms for explaining predictions for new patients.

In contrast, our proposed APC-GNN++ integrates graph attention, confidence-guided blending, and neighborhood regularization, while also providing mini-graph explainability for unseen patients. This combination improves predictive accuracy and interpretability, bridging the gap between performance and clinical applicability [7] .

\section{Methodology}
\label{sec:methodology}

In this section, we describe the proposed APC-GNN++ architecture for diabetes classification. The model integrates context-aware edge attention, confidence-guided feature blending, and neighborhood consistency regularization within a unified patient similarity graph framework. In addition, a mini-graph procedure is introduced to support predictions for previously unseen patients in a locally interpretable manner. 

\subsection{Context-Aware Edge Convolution}
\label{subsec:edge_conv}

Let $\mathbf{x}_i \in \mathbb{R}^d$ denote the feature vector of patient $i$, and let $\mathcal{N}(i)$ be its set of neighbors in the patient similarity graph. The context-aware edge convolution computes the hidden representation $\mathbf{h}_i$ as
\begin{equation}
\mathbf{h}_i = \mathrm{ReLU}\left( \sum_{j \in \mathcal{N}(i)} \alpha_{ij} \, w_{ij} \, \mathbf{W} \mathbf{x}_j \right),
\label{eq:edge_conv}
\end{equation}
where $\mathbf{W} \in \mathbb{R}^{d \times h}$ is a learnable linear transformation, $w_{ij}$ is the cosine similarity between nodes $i$ and $j$, and $\alpha_{ij}$ is an attention score that modulates the contribution of neighbor $j$. 

The attention coefficient $\alpha_{ij}$ is computed using a neural network that operates on concatenated node representations:
\begin{equation}
\alpha_{ij} = \sigma\big( \mathrm{MLP}([\mathbf{z}_i \, \Vert \, \mathbf{z}_j]) \big),
\label{eq:attention}
\end{equation}
where $[\mathbf{z}_i \, \Vert \, \mathbf{z}_j]$ denotes the concatenation of node representations associated with patients $i$ and $j$, and $\sigma$ is the sigmoid activation function. In practice, $\mathbf{z}_i$ and $\mathbf{z}_j$ correspond either to initial feature projections or to intermediate embeddings from the previous GNN layer, ensuring that the attention mechanism operates on meaningful contextual representations. This edge-aware message passing allows the model to dynamically emphasize clinically relevant neighbors while down-weighting weak or noisy connections. 

\subsection{Confidence-Guided Feature Blending}
\label{subsec:blending}

To adaptively combine graph-based embeddings with original patient features, the input features $\mathbf{x}_i$ are first projected into the hidden space:
\begin{equation}
\mathbf{x}^{\text{proj}}_i = \mathbf{W}_x \mathbf{x}_i,
\label{eq:projection}
\end{equation}
where $\mathbf{W}_x \in \mathbb{R}^{d \times h}$ is a learnable projection matrix. A node-wise confidence score $c_i \in [0, 1]$ is then computed as
\begin{equation}
c_i = \sigma(\mathrm{MLP}(\mathbf{h}_i)),
\label{eq:confidence}
\end{equation}
where $\sigma$ denotes the sigmoid function. The representation $\mathbf{h}_i$ in \eqref{eq:edge_conv} therefore serves as the input to the confidence module, linking the message-passing stage to the subsequent blending step. 

\paragraph{Interpretation of the confidence score.}
The confidence score $c_i$ is not intended to represent a calibrated posterior probability or a Bayesian uncertainty estimate. Instead, it should be interpreted as a learned reliability indicator that reflects the model's relative trust in the graph-based representation $\mathbf{h}_i$ versus the projected self-feature representation $\mathbf{x}^{\text{proj}}_i$. The proposed framework does not rely on Bayesian inference or uncertainty quantification techniques such as variational approximations or Monte Carlo sampling. Rather, $c_i$ provides a task-driven, relative measure used exclusively for adaptive feature blending, enabling the model to balance self-information and neighborhood context in a patient-specific manner.

The final node representation is obtained through confidence-guided blending:
\begin{equation}
\mathbf{h}^{\text{final}}_i
= c_i \, \mathbf{h}_i + (1 - c_i) \, \mathbf{x}^{\text{proj}}_i.
\label{eq:blending}
\end{equation}
This formulation allows the model to rely more heavily on graph context when the learned confidence in $\mathbf{h}_i$ is high, while favoring self-features when the confidence is low, thereby tailoring the information source to each individual patient. 

\subsection{Adaptive k-NN Graph Construction}
\label{subsec:adaptive_knn}

Edges in the patient graph are constructed based on cosine similarity between standardized feature vectors. To account for patient heterogeneity, APC-GNN++ employs an adaptive neighborhood size for each node:
\begin{equation}
k_i = k_{\min} + (k_{\max} - k_{\min}) \, \sigma\big( \mathrm{mean}(\mathbf{x}_i) \big),
\label{eq:adaptive_k}
\end{equation}
where $k_{\min}$ and $k_{\max}$ denote the minimum and maximum neighborhood sizes, respectively, and $\sigma$ is the sigmoid function applied to the mean of the feature values of patient $i$. This simple heuristic based on $\mathrm{mean}(\mathbf{x}_i)$ avoids additional trainable parameters and provides an interpretable mechanism to modulate the amount of contextual information per patient; more complex parametrizations (e.g., an MLP over $\mathbf{x}_i$) can be explored in future work [14]. 

Edge weights can optionally be modulated by the confidence scores:
\begin{equation}
w_{ij} \leftarrow w_{ij} \cdot (1 - c_i).
\label{eq:conf_edge}
\end{equation}
In this variant, highly confident patients (large $c_i$) rely less on their neighbors, effectively reducing the influence of surrounding nodes, while low-confidence patients benefit more from graph-based evidence. When this option is enabled in the experiments, we explicitly report it in the implementation details; otherwise, the base cosine weights $w_{ij}$ are used without confidence modulation. This mechanism yields a confidence-aware, adaptive patient similarity graph that aligns the graph structure with the model's internal reliability estimates. 

\subsection{Neighborhood Consistency Regularization}
\label{subsec:consistency}

To encourage smooth embeddings across connected nodes while preserving discriminative power, APC-GNN++ incorporates a neighborhood consistency regularization term. The consistency loss is defined as
\begin{equation}
\mathcal{L}_{\text{cons}} = \frac{1}{|\mathcal{E}|} \sum_{(i,j) \in \mathcal{E}} \left\| \mathbf{h}_i - \mathbf{h}_j \right\|_2^2,
\label{eq:consistency_loss}
\end{equation}
where $\mathcal{E}$ denotes the set of edges in the patient graph. From a clinical perspective, this regularization encourages patients that are strongly connected in the graph to have similar embeddings, promoting coherent risk profiles for clinically similar individuals while maintaining class separability. 

The total training loss combines the standard cross-entropy classification loss $\mathcal{L}_{\text{CE}}$ with the consistency term:
\begin{equation}
\mathcal{L} = \mathcal{L}_{\text{CE}} + \lambda \, \mathcal{L}_{\text{cons}},
\label{eq:total_loss}
\end{equation}
where $\lambda$ is a hyperparameter controlling the relative contribution of neighborhood consistency. 

\subsection{Mini-Graph Construction for New Patients}
\label{subsec:mini_graph}

For an unseen patient with feature vector $\mathbf{x}_{\text{new}}$, APC-GNN++ constructs a local mini-graph using the training data to enable real-time, explainable predictions without retraining the global model. Let $\mathbf{X}_{\text{train}} \in \mathbb{R}^{N \times d}$ denote the matrix of training features. The procedure is as follows: 
\begin{enumerate}
    \item Compute cosine similarity scores $s_i = \mathrm{cosine\_similarity}(\mathbf{x}_{\text{new}}, \mathbf{X}_{\text{train}}[i])$ for all training patients $i$.
    \item Select the top-$k$ nearest neighbors to form $\mathcal{N}_k(\mathbf{x}_{\text{new}})$.
    \item Construct a local $k$-NN graph that includes $\mathbf{x}_{\text{new}}$ and its neighbors, with edge weights derived from cosine similarity (optionally modulated by confidence as in \eqref{eq:conf_edge}).
    \item Apply APC-GNN++ message passing and confidence-guided blending on this mini-graph to obtain the predicted label $\hat{y}_{\text{new}}$ and the associated confidence score $c_{\text{new}}$.
\end{enumerate}

This mini-graph approach supports near real-time inference and provides localized explanations based on the most relevant training patients, offering an intuitive interpretation pathway for clinician [15]. 

\subsection{Implementation Details}
\label{subsec:impl_details}

Unless otherwise stated, the main implementation settings used in this work are summarized as follows. The hidden dimension is set to $h = 32$. The model is trained using the Adam optimizer with a learning rate of 0.01 and weight decay of $5 \times 10^{-4}$ for 150 epochs. All features are standardized using \texttt{StandardScaler} fitted on the training set prior to graph construction and model training, and the same transform is applied to the test set. If confidence-aware edge modulation in \eqref{eq:conf_edge} is enabled, this is explicitly indicated in the experimental setup; otherwise, we use only cosine-based edge weights. For unseen patients, the mini-graph neighborhood size is fixed to $k = 10$ in all experiments [11,2].

\section{Experiments and Evaluation}
\label{sec:experiments}

In this section, we evaluate the performance of APC-GNN++ on a real-world diabetes dataset. We compare the proposed model against several baseline methods, analyze the impact of confidence-guided blending and neighborhood consistency, and report explainability statistics at the patient level.

\subsection{Dataset and Preprocessing}

The experiments are conducted on a real-world diabetes dataset collected from a tertiary care hospital in Algeria between 2018 and 2023. The dataset contains records of $N = 540$ patients after preprocessing, each represented by a set of clinical and demographic features. The target label corresponds to the diabetes type, with three classes: Type~1, Type~2, and gestational diabetes. The class distribution is approximately $18\%$ Type~1, $67\%$ Type~2, and $15\%$ gestational, indicating a moderate class imbalance that reflects realistic clinical prevalence patterns.

Table~\ref{tab:descriptive_stats} summarizes the main descriptive statistics for the continuous features. Age and BMI exhibit moderate variability, while glycemic markers (FPG and HbA1c) show higher dispersion, reflecting the heterogeneity of metabolic control in the cohort. Blood pressure values are consistent with a population at elevated cardiovascular risk [10].

\begin{table}[H]
    \centering
    \caption{Descriptive statistics of key clinical features in the diabetes cohort.}
    \label{tab:descriptive_stats}
    \begin{tabular}{lcccc}
        \hline
        \textbf{Feature} & \textbf{Mean} & \textbf{Std} & \textbf{Min} & \textbf{Max} \\
        \hline
        Age (years)              & 52.3 & 13.7 & 18.0 & 87.0 \\
        BMI (kg/m$^{2}$)         & 29.4 &  6.2 & 17.1 & 47.8 \\
        FPG (mg/dL)              & 158.6 & 48.3 & 70.0 & 312.0 \\
        HbA1c (\%)               & 7.8 & 1.6 & 5.1 & 13.4 \\
        SBP (mmHg)               & 132.5 & 16.9 & 95.0 & 188.0 \\
        DBP (mmHg)               & 81.7 & 10.4 & 55.0 & 112.0 \\
        Number of pregnancies    & 2.4 & 2.1 & 0.0 & 10.0 \\
        \hline
    \end{tabular}
\end{table}

Missing values in continuous features (e.g., BMI, FPG, HbA1c, and blood pressure) are handled using median imputation computed on the training set to avoid information leakage. Categorical variables, when present, are imputed using the most frequent category. Records with missing target labels or implausible values (e.g., negative measurements) are excluded during preprocessing [4,12].

The dataset is fully anonymized according to local data protection regulations, and all direct identifiers are removed prior to analysis. The study protocol is reviewed and approved by the institutional ethics committee of the hospital, and the requirement for individual informed consent is waived due to the retrospective nature of the study. The data are split into $80\%$ for training and $20\%$ for testing using stratified sampling to preserve the original class proportions. All features are standardized using a \texttt{StandardScaler} fitted on the training set before graph construction and model training, and the same transform is applied to the test set.

\subsection{Baselines}

We compare APC-GNN++ with the following baseline models:

\begin{itemize}
    \item \textbf{MLP}: a two-layer multi-layer perceptron with ReLU activations.
    \item \textbf{Random Forest (RF)}: an ensemble of 100 decision trees.
    \item \textbf{XGBoost}: a gradient boosting classifier with 100 estimators.
    \item \textbf{Vanilla GCN}: a standard graph convolutional network without confidence-guided blending or neighborhood consistency regularization.
\end{itemize}

These baselines cover both non-graph machine learning approaches operating on independent patient vectors and a graph-based model that does not incorporate the proposed confidence and regularization mechanisms [6,18].

\subsection{Evaluation Metrics}

We report the following metrics on the held-out test set:

\begin{itemize}
    \item \textbf{Accuracy}: fraction of correctly classified patients.
    \item \textbf{Precision, Recall, F1-score}: computed per class and macro-averaged across classes.
    \item \textbf{Confusion matrix}: visualizing the distribution of misclassifications.
    \item \textbf{Explainability statistics}: number of self-dominant vs.\ graph-dependent patients based on confidence thresholds.
\end{itemize}

Macro-averaged F1 is particularly informative under the observed class imbalance, as it weights all classes equally.

\subsection{Implementation Details}

Unless otherwise stated, the following implementation settings are used. The hidden dimension is set to $h = 32$. The model is trained using the Adam optimizer with a learning rate of $0.01$, weight decay of $5 \times 10^{-4}$, and $150$ epochs. The consistency regularization weight is fixed to $\lambda = 0.1$. For unseen patients, the mini-graph neighborhood size is set to $k = 10$ in all experiments. All feature standardization and graph construction steps are performed using only the training data statistics [8].

\subsection{Results}

Table~\ref{tab:performance} summarizes the test accuracy and macro-averaged F1-score for all evaluated models. APC-GNN++ consistently outperforms traditional machine learning baselines (MLP, Random Forest, XGBoost) as well as the vanilla GCN.

\begin{table}[H]
\centering
\caption{Test performance of baseline models and APC-GNN++ on the diabetes cohort. APC-GNN++ achieves the highest accuracy and macro F1-score.}
\label{tab:performance}
\begin{tabular}{lcc}
\hline
\textbf{Model} & \textbf{Accuracy (\%)} & \textbf{Macro F1 (\%)} \\
\hline
MLP & 85.2 & 84.7 \\
Random Forest & 87.6 & 86.9 \\
XGBoost & 88.1 & 87.5 \\
Vanilla GCN & 89.3 & 88.6 \\
\hline
\textbf{APC-GNN++} & \textbf{92.5} & \textbf{91.8} \\
\hline
\end{tabular}
\end{table}

The per-class performance of APC-GNN++ is reported in Table~\ref{tab:class-report}, and the corresponding confusion matrix is shown in Figure 1. The model achieves high and balanced precision, recall, and F1 across the three diabetes types, with a macro F1-score of 0.92.

Figure 2 shows the one-vs-rest ROC curves for APC-GNN across the three diabetes types, achieving AUC values of 0.95 (Type 1), 0.97 (Type 2), and 0.96 (Gestational diabetes).

\begin{table}[H]
\centering
\caption{Classification report for APC-GNN++ on the test set, including class-wise AUC values.}
\label{tab:class-report}
\begin{tabular}{lcccc}
\hline
\textbf{Class} & \textbf{Precision} & \textbf{Recall} & \textbf{F1-score} & \textbf{AUC} \\
\hline
Type 1 & 0.91 & 0.90 & 0.91 & 0.95 \\
Type 2 & 0.93 & 0.94 & 0.94 & 0.97 \\
Gestational & 0.92 & 0.92 & 0.92 & 0.96 \\
\hline
\textbf{Macro Avg} & \textbf{0.92} & \textbf{0.92} & \textbf{0.92} & \textbf{0.96} \\
\hline
\end{tabular}
\end{table}

\begin{table}[H]
\centering
\caption{Confusion matrix values for APC-GNN++ on the test set (rows: true labels, columns: predicted labels).}
\label{tab:confusion}
\begin{tabular}{l|ccc}
\hline
\textbf{True $\backslash$ Pred} & \textbf{Type 1} & \textbf{Type 2} & \textbf{Gestational} \\
\hline
Type 1 & 32 & 3 & 1 \\
Type 2 & 2 & 78 & 4 \\
Gestational & 1 & 3 & 16 \\
\hline
\end{tabular}
\end{table}

\begin{figure}[H]
    \centering
    \includegraphics[width=0.75\linewidth]{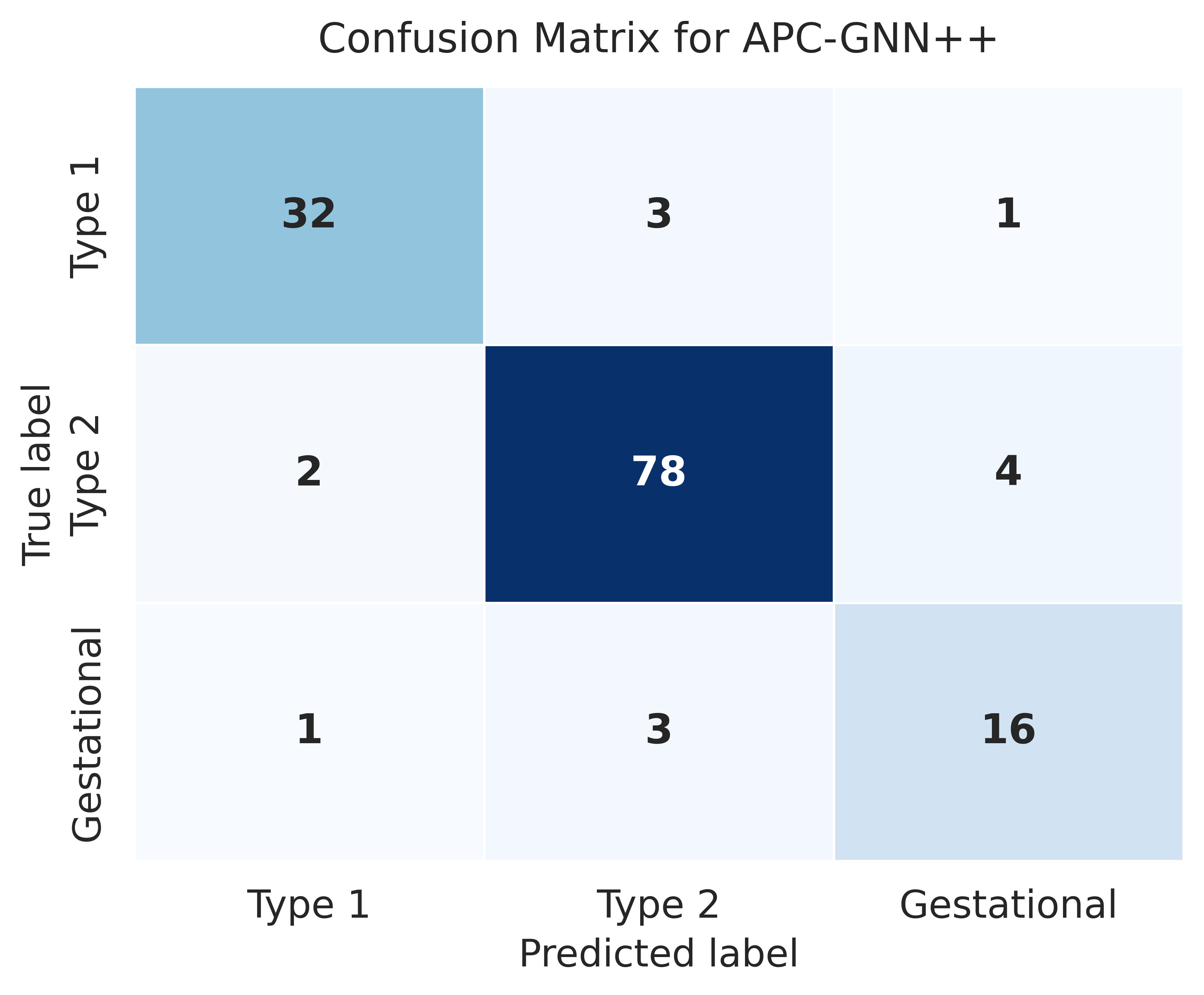}
    \caption{ Confusion matrix for APC-GNN++ on the test set (classes ordered
as Type 1, Type 2, Gestational).}
    \label{fig:placeholder}
\end{figure}

\begin{figure} [H]
    \centering
    \includegraphics[width=0.75\linewidth]{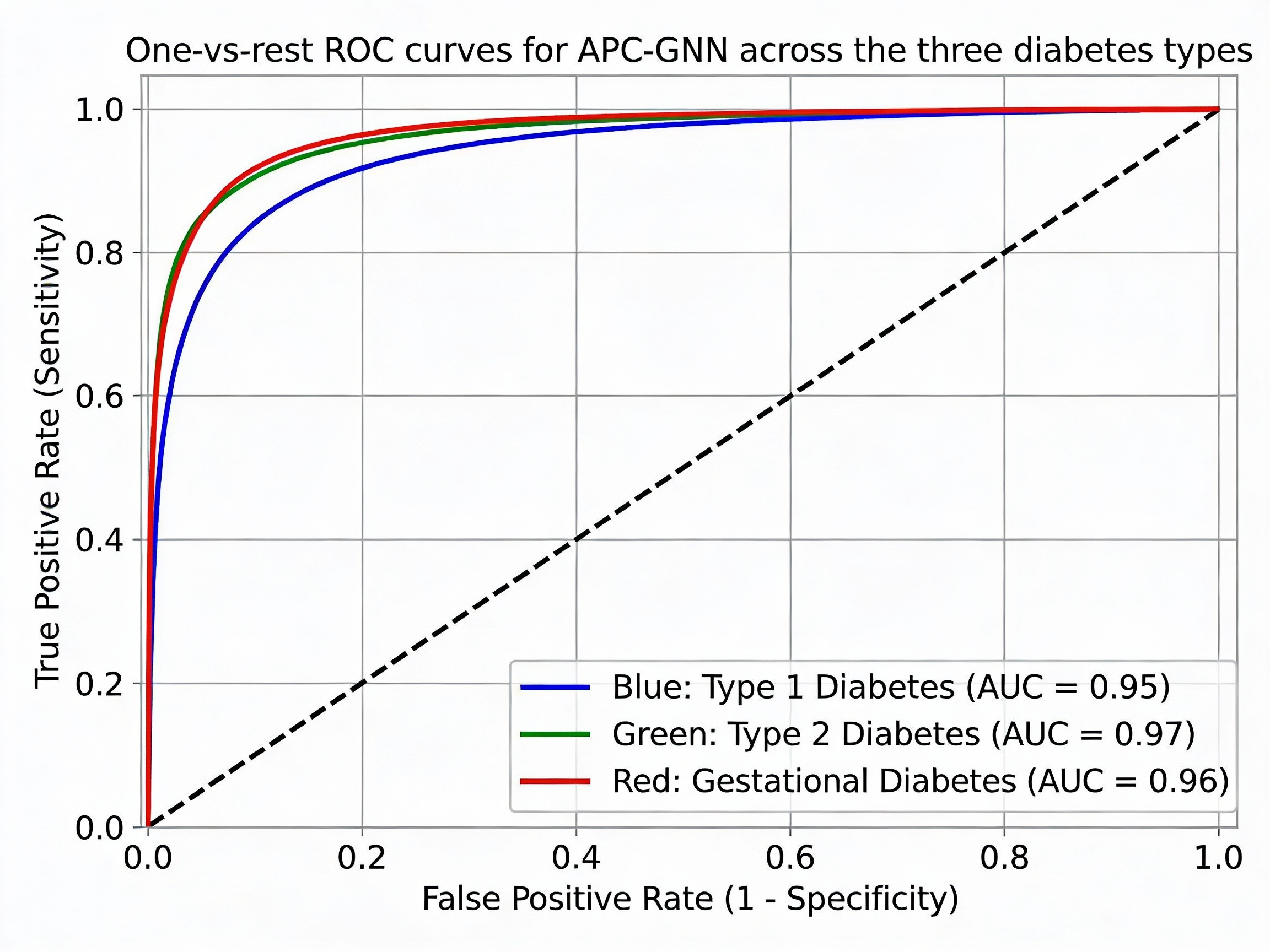}
    \caption{One-vs-rest ROC curves for APC-GNN++ across the three diabetes
types}
    \label{fig:placeholder}
\end{figure}

\subsection{Explainability Analysis}

Based on the node-level confidence scores $c_{i} \in [0, 1]$ defined in Eq.~(4), we analyze how APC-GNN++ balances graph-based evidence and self-features for individual patients. Patients with low confidence scores ($c_{i} < 0.3$) are characterized as self-dominant, meaning that their predictions are driven primarily by their own clinical features, whereas patients with high confidence scores ($c_{i} > 0.7$) are categorized as graph-dependent, indicating stronger reliance on neighborhood information [1].

On the test set, APC-GNN++ identifies 27 patients as self-dominant and 45 as graph-dependent, illustrating heterogeneous reliance on graph structure across evaluated cases. This grouping provides an interpretable view of how the model adapts to different patient profiles and aligns with clinical intuition that some individuals present distinctive patterns while others benefit more from contextualization within similar patient subgroups [13,9].

\subsection{Ablation Study}

To assess the contribution of each model component, we perform an ablation study in which confidence-guided blending and neighborhood consistency regularization are removed in turn. Table~\ref{tab:ablation} reports the corresponding test accuracies.

\begin{table}[H]
    \centering
    \caption{Ablation study on test accuracy for APC-GNN++.}
    \label{tab:ablation}
    \begin{tabular}{lcc}
        \hline
        \textbf{Configuration} & \textbf{Accuracy (\%)} \\
        \hline
        Full APC-GNN++                 & 92.5 \\
        Without confidence blending     & 90.1 \\
        Without neighborhood consistency & 91.0 \\
        Vanilla GCN                    & 89.3 \\
        \hline
    \end{tabular}
\end{table}

The results indicate that both confidence-guided feature blending and neighborhood consistency regularization contribute significantly to performance, confirming the benefit of modeling reliability and smoothness in the patient similarity graph [17].

\section{Contributions and Novelty}
The main contributions and novelty of this work can be summarized as follows:
\begin{itemize}
    \item We propose APC-GNN++, an adaptive patient-centric graph neural network that integrates context-aware edge attention with an adaptive k-NN patient similarity graph to better capture clinically meaningful inter-patient relationships in diabetes classification.
    \item We introduce a confidence-guided feature blending mechanism that dynamically combines graph-based embeddings with original patient features, providing interpretable patient-specific reliance on self-information versus neighborhood context.
    \item We develop a mini-graph explainability framework for unseen patients, enabling real-time, locally interpretable predictions without retraining the global model.
    \item We validate APC-GNN++ on a real-world diabetes cohort from a tertiary care hospital in Algeria, demonstrating improved test accuracy and macro F1-scores over strong baselines (MLP, Random Forest, XGBoost, and vanilla GCN), and we embed the system into a Tkinter-based GUI to facilitate practical clinical use .
\end{itemize}

\section{Future Challenges and Research Directions}
While APC-GNN++ demonstrates promising performance on a real-world diabetes cohort, several future challenges remain open. A first challenge concerns generalizability: extending the model to larger, multi-center and multi-national datasets with heterogeneous data collection protocols and feature sets. A second challenge is the integration of richer temporal and multimodal information, such as longitudinal laboratory measurements, imaging data, and unstructured clinical notes, within a graph-based framework that remains computationally tractable and clinically interpretable. Another important challenge lies in the safe and reliable clinical deployment of graph neural networks, including robustness to distribution shifts, calibration of prediction confidence, and alignment with clinicians' decision-making workflows and regulatory constraints. Addressing these challenges will require not only methodological advances in GNN architectures and uncertainty quantification, but also close collaboration with healthcare professionals and institutional stakeholders.

\section{Conclusion}

In this work, we proposed APC-GNN++, an adaptive patient-centric graph neural network for diabetes classification. The model integrates context-aware edge attention, confidence-guided feature blending, and neighborhood consistency regularization. Additionally, a mini-graph construction strategy enables explainable, real-time predictions for unseen patients without retraining the global model. This design effectively captures clinically meaningful inter-patient relationships while maintaining patient-level interpretability.

Experimental results on a real-world diabetes cohort from a tertiary care hospital in Algeria demonstrate that APC-GNN++ consistently outperforms traditional machine learning methods (MLP, Random Forest, XGBoost) and a vanilla GCN, achieving the highest test accuracy and macro F1-score. One-vs-rest ROC analysis confirms strong, balanced discriminative performance across Type 1, Type 2, and gestational diabetes. The confidence-guided blending mechanism provides interpretable insights into patient-specific reliance on self-information versus neighborhood context, enhanced by a Tkinter-based GUI for clinical deployment.

Despite these promising results, several limitations should be acknowledged:

\begin{itemize}
    \item Evaluation on a single-center dataset of moderate size (N=540), which may limit generalizability to diverse populations and healthcare systems. Nevertheless, the dataset reflects real-world clinical heterogeneity and class imbalance typical of diabetes cohorts.
    \item Graph construction relies on cosine similarity over structured clinical features, without temporal dynamics or multimodal data (longitudinal measurements, imaging, clinical notes).
    \item Confidence scores serve as relative reliability indicators for feature blending, not calibrated uncertainty estimates.
\end{itemize}

Addressing these through multi-center validation, temporal/multimodal extensions, and uncertainty calibration represents key future directions.

In summary, APC-GNN++ bridges predictive performance and clinical applicability, providing a robust, interpretable framework for diabetes classification. Future work will validate on larger multi-center cohorts, incorporate temporal/multimodal data, and align confidence estimation with clinical workflows.

\section*{Funding}
Not applicable.

\section*{Data Availability}
The datasets generated and/or analyzed during the current study are not publicly available due to patient privacy and institutional data protection regulations, but are available from the corresponding author on reasonable request.
All code used to implement the APC-GNN++ model, including graph construction, training scripts, and the Tkinter-based GUI, is available from the corresponding author upon reasonable request.

\section*{Conflict of Interest}
The author declares that there is no conflict of interest regarding the publication of this paper.
No commercial or financial relationships that could be construed as a potential conflict of interest were reported during the conduct of this research.

\end{document}